\documentclass[11pt]{article}
\usepackage{times}
\usepackage{amssymb,amsmath}
\usepackage{graphicx}
\usepackage{microtype}
\usepackage[square]{natbib}
\usepackage{caption}
\usepackage[hidelinks, colorlinks, linkcolor={red},citecolor={blue}, urlcolor={blue}, backref=section]{hyperref}
\usepackage[left=1in, right=1in, top=1.0in, bottom=1.0in]{geometry}
\usepackage[capitalise]{cleveref}
\usepackage{authblk}
\usepackage{blindtext}
\usepackage{subcaption}

\title{3D Fusion of Infrared Images with Dense RGB Reconstruction from Multiple Views
- with Application to Fire-fighting Robots}
\date{June 30, 2013}

\author[1]{Yuncong Chen}
\author[2]{Will Warren}
\affil[1]{Department of Computer Science and Engineering, UCSD}
\affil[2]{Department of Mechanical and Aerospace Engineering, UCSD}

\begin{document}

\maketitle

\begin{center}
Abstract: This project integrates infrared and RGB imagery to produce dense 3D environment models reconstructed from multiple views. The resulting 3D map contains both thermal and RGB information which can be used in robotic fire-fighting applications to identify victims and active fire areas.

\end{center}

\section{Introduction}

Modern robotic systems are increasingly practical for urgent, data-oriented challenges such as firefighting, search and rescue operations, environmental monitoring and defense. Advancements in sensor technology, robot versatility and machine perception are accelerating the development of compact robots which can interact with the environment in intelligent ways, completing tasks of relative great complexity in individual and synchronized efforts.

These new technologies offer an opportunity to transform the way in which modern firefighting teams address life threatening situations. Deploying sensor-laden unmanned vehicles to sweep through high-risk urban environments would yield incredible gains in situational awareness and remove immediate threats to firefighters' lives. A properly equipped robot could act as a firefighter's eyes and ears before human lives are unnecessarily placed at risk, and a modular payload could allow for a plethora of sensing information hitherto unattainable. The UCSD Coordinated Robotics Lab is focused on the design, implementation, and testing of semi-autonomous sensor-laden vehicles that are capable of traversing and mapping urban environments while identifying survivors and active fire areas.


\section{Fire Fighting Robot}

\begin{figure}[h]
	\begin{subfigure}[b]{.4\linewidth}
	\centering
	\includegraphics[width=\linewidth]{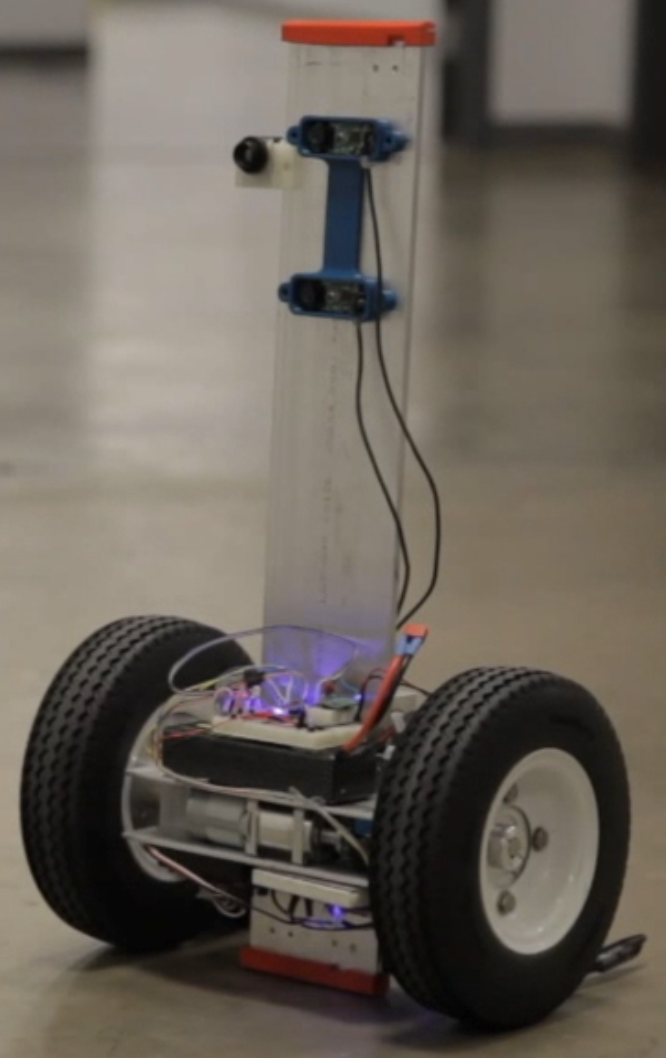}
	\label{fig:robot}
	\end{subfigure}%
        ~ 
    \begin{subfigure}[b]{.45\linewidth}
    \centering
    \includegraphics[width=\linewidth]{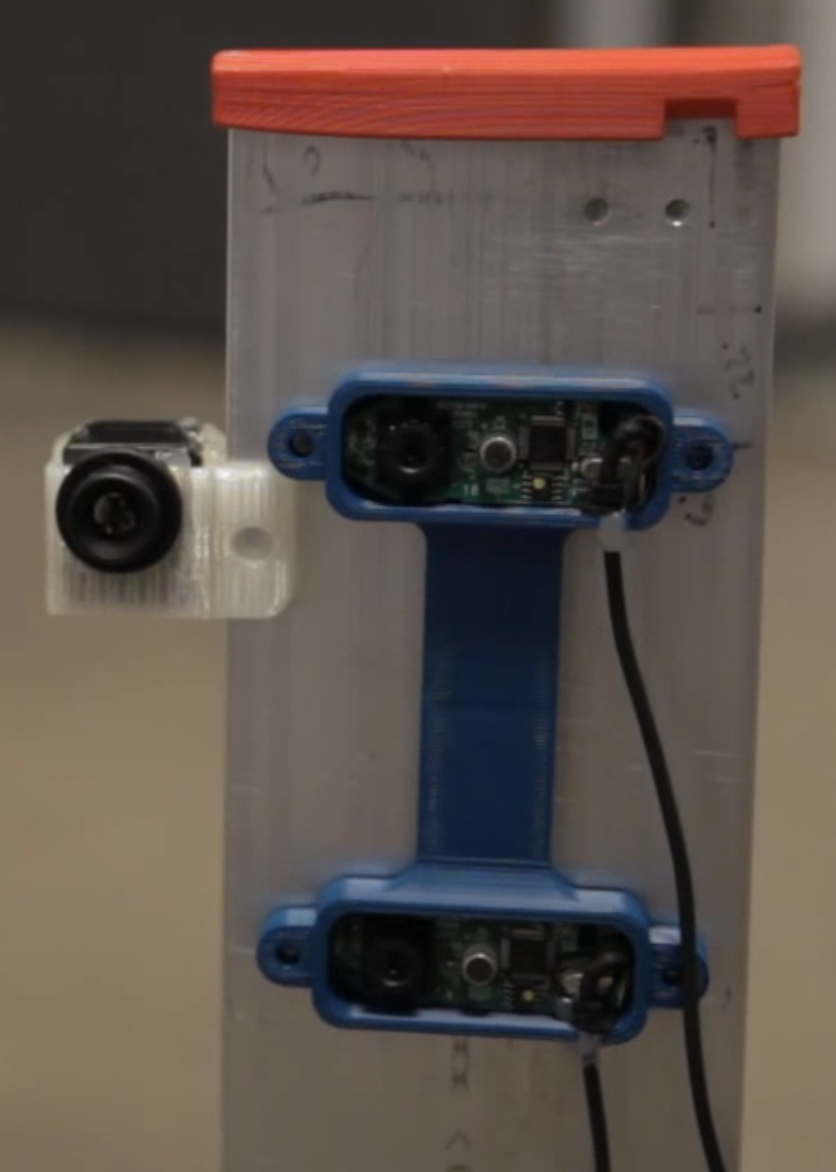}
    \label{fig:sensor}
    \end{subfigure}
   \caption{(a) the FFR. (b) on-board sensors: a Tamarisk 320 infrared imager and two Logitech webcams}\label{fig:robot_sensor}
\end{figure}

FFR is a ruggedized, physically robust scout vehicle whose purpose is to aid and assist first responders in structure fires, surveillance, and search and rescue. FFR traverses terrain using a Segway like balancing motion which raises the sensor package to a higher vantage point. The robot is unique in that it has a center leg that is actuated up and down, allowing it to climb stairs and overcome large obstacles. The robot's vision system consists of dual Logitech C320 web cameras coupled with a Tamarisk 320 infrared imager (Figure \ref{fig:robot_sensor}). This configuration is ideal because it meets the strict weight and form factor constraints inherent to mobile robot applications.

\section{Camera Calibration}

Calibration is used to resolve the intrinsic parameters of each camera (i.e. focal length, principle point), as well as their relative position and orientation with respect to each other. These parameters establish the projective relationship between a 3D point in the world coordinate and the corresponding 2D point in each camera's image plane. Such a relationship is the basis for the following multiple view reconstruction and back-projection of thermal image data.

\begin{figure}[h]
        \centering
                \begin{subfigure}[b]{0.32\linewidth}
                \centering
  \includegraphics[width=\linewidth]{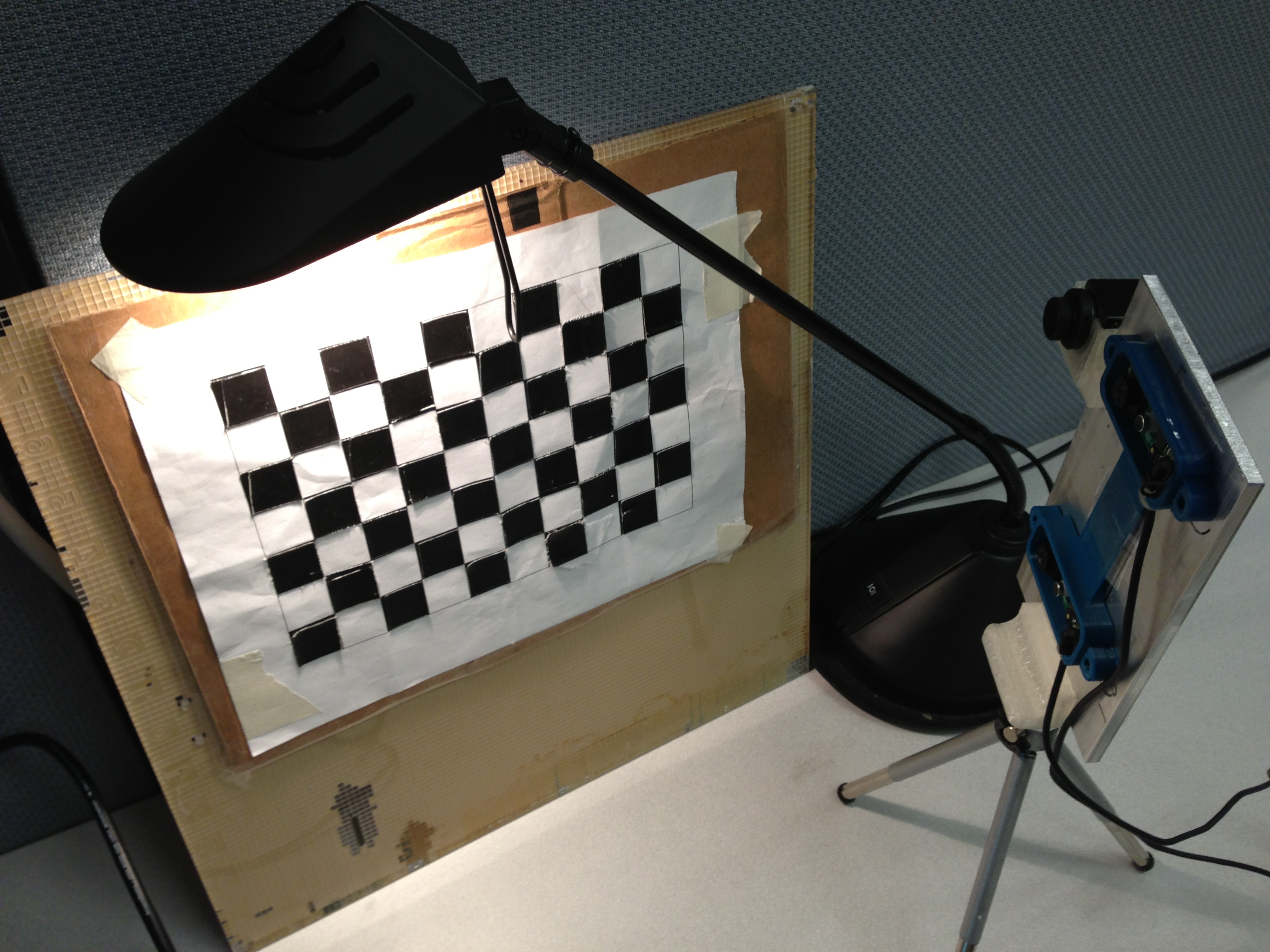}
  \label{fig:setup}
        \end{subfigure}%

        \begin{subfigure}[b]{0.32\linewidth}
                \centering
  \includegraphics[width=\linewidth]{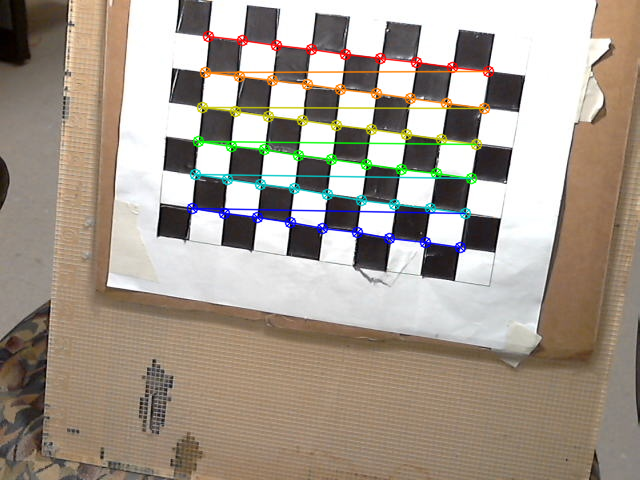}
  \label{fig:calib_rgb}
        \end{subfigure}%
        ~ 
        \begin{subfigure}[b]{0.32\linewidth}
                \centering
  \includegraphics[width=\linewidth]{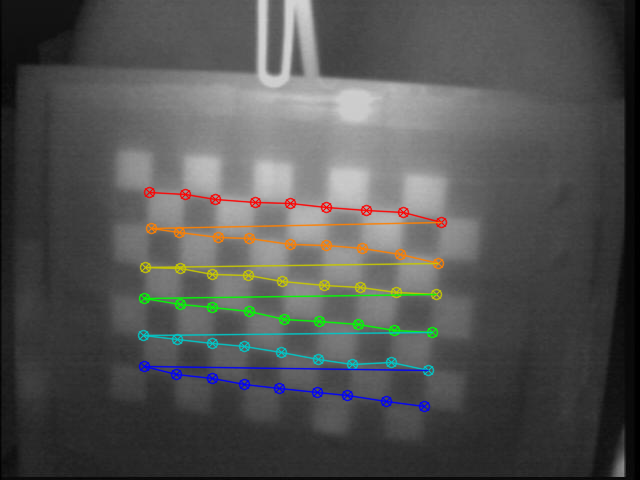}
  \label{fig:calib_thermal}
  \end{subfigure}
   \caption{Top: the setup for camera calibration. Bottom: the calibration pattern in (left) RGB camera and (right) thermal camera. Colored circles represent detected corners.}\label{fig:calib}
\end{figure}

A classical approach to camera calibration uses a ``calibration object'' which has a simple structure with known dimensions such that features (i.e. corners), as well as their world coordinates, may easily be detected within imagery. The mapping between features' image location and world coordinates are then used to infer the camera parameters. In our experiments, a planar checkerboard pattern was used. The corners within the pattern were detected and matched using OpenCV\footnote{http://opencv.org}. Thermally conductive rubber tape was placed over the checkerboard's black squares, producing a large temperature gradient at each corner when placed under a heat lamp. The setup and sample calibration images are shown in Figure \ref{fig:calib}.

%

\section{Visual Structure from Motion in RGB}


As the robot navigates through a given structure it captures a sequence of images for both the infrared and RGB cameras. Three dimensional depth data may be resolved from the set of RGB images using visual structure from motion, producing a dense point cloud with RGB textures. For the purposes of this experiment, open source VisualSFM software \cite{siftgpu07wu} \cite{Wu11multicorebundle} \cite{vsfm} was used.



\begin{figure}[h]
        \centering
        \begin{subfigure}[b]{0.48\linewidth}
            \centering
  			\includegraphics[width=\linewidth]{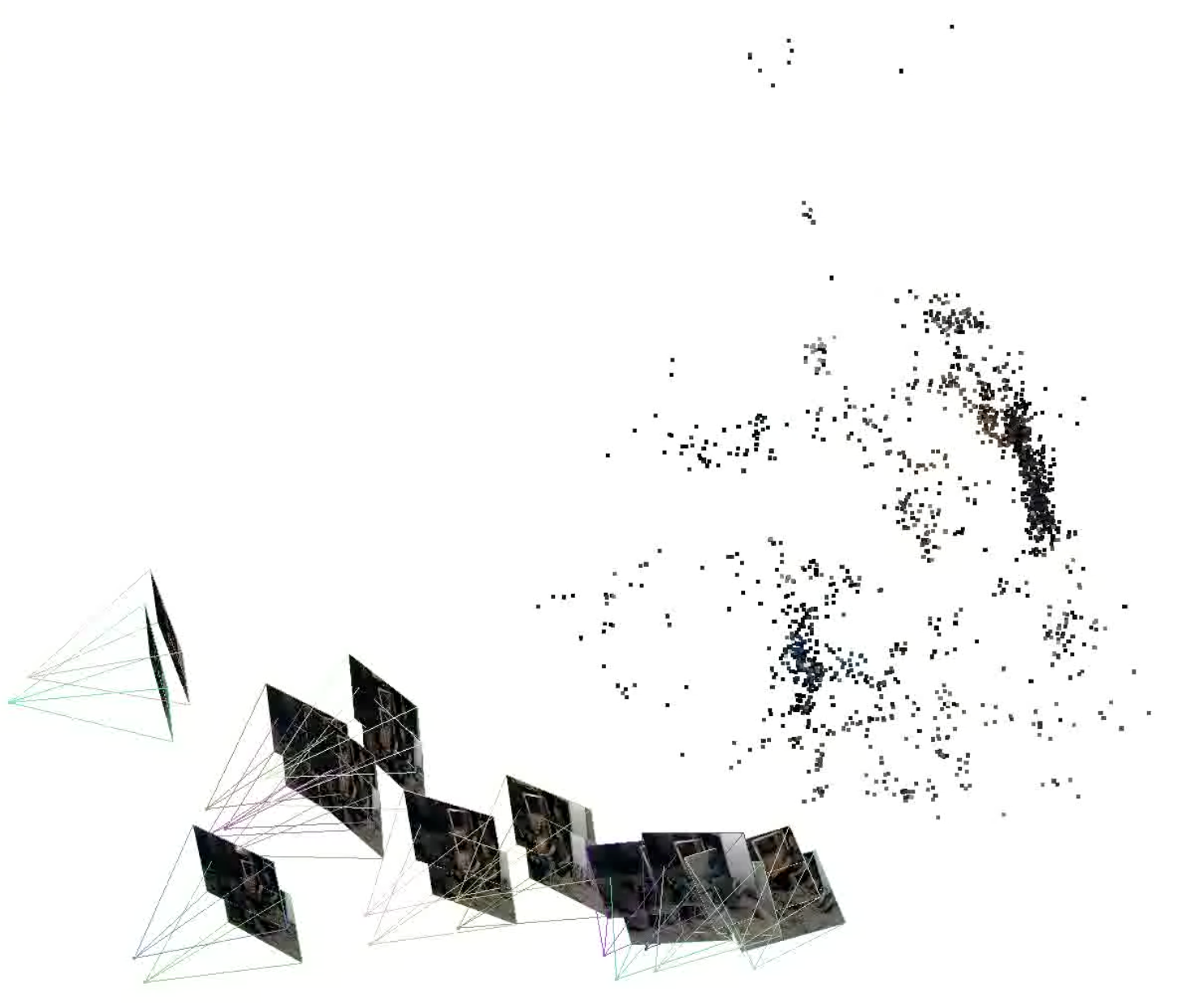}
        \end{subfigure}%
        ~ 
        \begin{subfigure}[b]{0.48\linewidth}
                \centering
  \includegraphics[width=\linewidth]{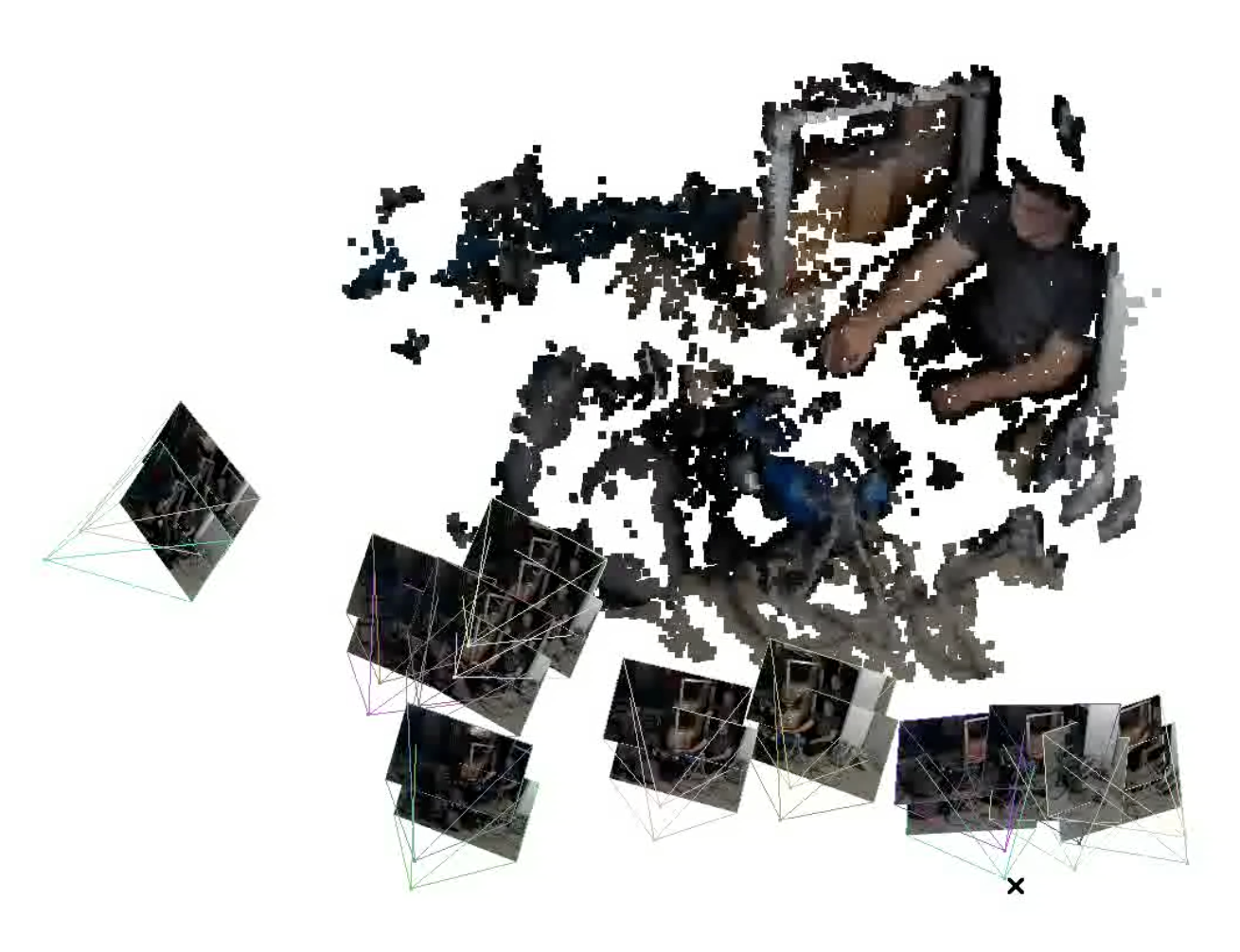}
  \end{subfigure}
   \caption{(a) a 3D point cloud of sparse keypoints in the scene, and the image planes of cameras at all frames. (b) the dense point cloud produced using PMVS/CMVS}\label{fig:cloud}
\end{figure}

This software detects keypoints in all images and compares them within a window of neighboring frames. Matched keypoints are grouped and regarded as measurements of the same point within the 3D scene. A global optimization procedure is then carried out to minimize the discrepancy between the actual measurements and the theoretical measurements computed from estimated camera pose (i.e. camera center position and orientation) for each frame and estimated 3D location for each keypoint in the scene. This procedure, referred to as bundle adjustment, results in a sparse cloud of points in the 3D world space, as well as a camera pose relative to the point cloud for each corresponding frame. Furukawa's PMVS/CMVS method \cite{Furu:2010:PMVS} \cite{Furu:2010} is used to expand the sparse cloud into a densely packed cloud of points (Figure \ref{fig:cloud}).

In fact, VisualSFM works even with images taken from a single RGB camera. The reason that we use stereo cameras is two-fold. Clearly, two cameras provide more complete coverage of the scene, collecting twice the measurements in the same time, but more importantly, a calibrated stereo rig determines the scale of the reconstructed point cloud. In fact, the reconstruction from VisualSFM is only up to a similarity transformation (i.e. rotation, translation and scaling). In order to project the reconstructed point cloud back to the calibrated thermal camera, we need to determine the true scale of the cloud. The most naive way to eliminate the scale ambiguity is to find Ground Control Points in the scene (points that have known world coordinates, or point pairs that have known distances), which in most cases are not possible. Our method however, does not rely on any external information. By comparing the true (known) baseline length (distance between two camera centers) of the stereo cameras with the estimated baseline lengths in the reconstructed point cloud, we can robustly calculate the scaling factor. The estimated baseline length is easily obtained from the estimated camera centers of the stereo pair provided by VisualSFM at each frame.

\section{Infrared Image Integration}

Given the dense cloud of 3D points, we project them to the thermal cameras, and find their thermal values from the thermal images.

Since the relative pose of the thermal camera with respect to the RGB cameras has been determined via calibration, we can find the pose of the thermal camera with respect to the point cloud. VisualSFM also gives information about which points are visible by the cameras at which frame. Using this information, each 3D point in the cloud is projected onto all thermal cameras from which this point is visible. The corresponding intensity values for this point are averaged and assigned to the point. This constitutes the 3D thermal cloud.

\section{Results}

We tested our system with persons sitting or lying in cluttered or dark environments, simulating the victim of a structure fire. As shown in Figure \ref{fig:will} and \ref{fig:garage}, the victim's image shows a significant contrast with the background. The accuracy of the thermal layer projection can be observed from the good alignment between the bright region in the thermal model and the contours of person in RGB point clouds. We also tested our approach on a standing person and racks of hot machines in a university server room. The thermal model of the standing person shows low thermal values at the precise locations of the backpack straps. The heat from the server machine is apparent in the thermal model and shows good contrast with the steel racks around it.

\begin{center}
\begin{figure}
  \includegraphics[width=\linewidth]{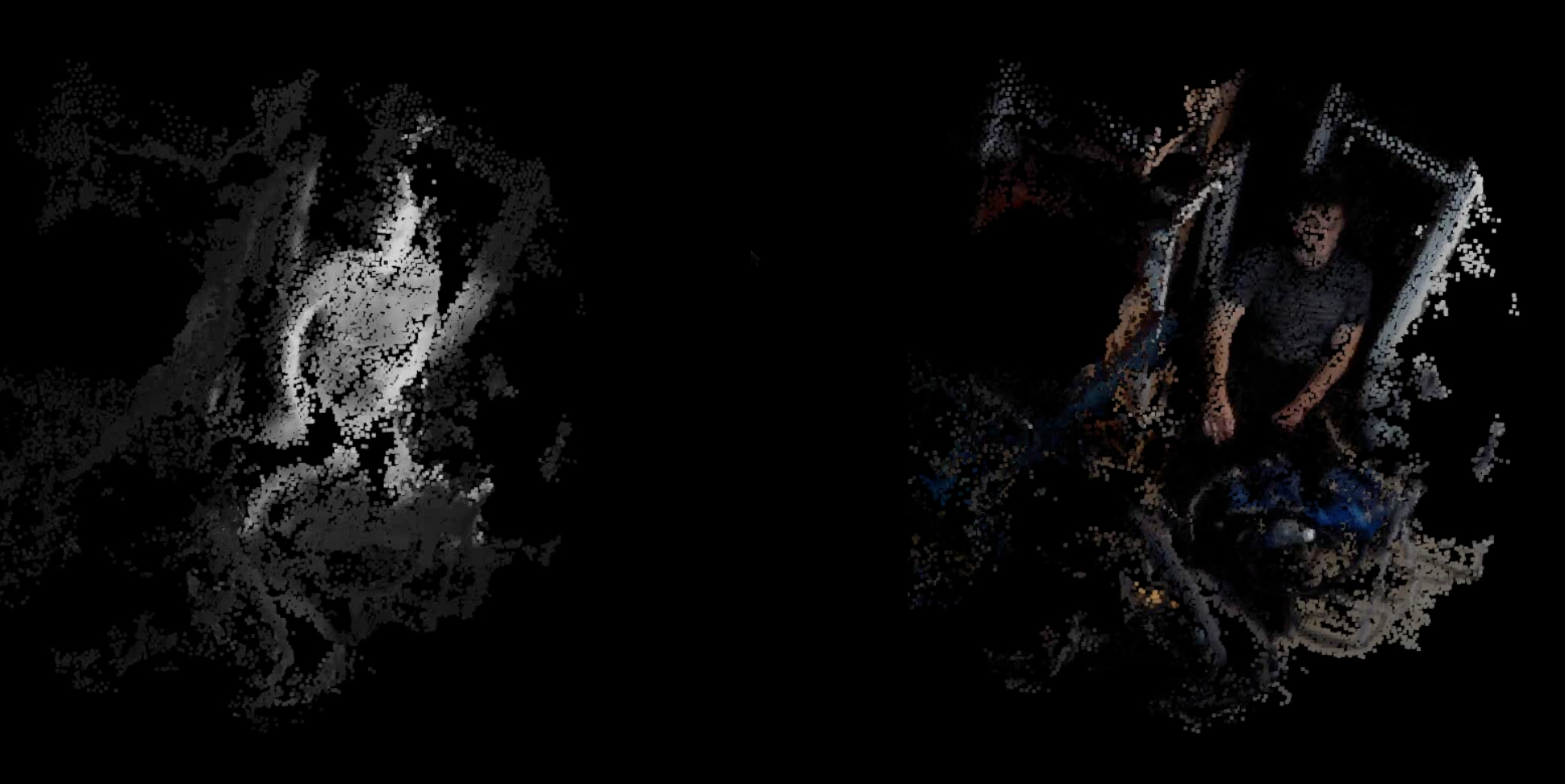}
   \caption{Victim sitting at a cluttered warehouse corner}
  \label{fig:will}
\end{figure}

\begin{figure}
  \includegraphics[width=\linewidth]{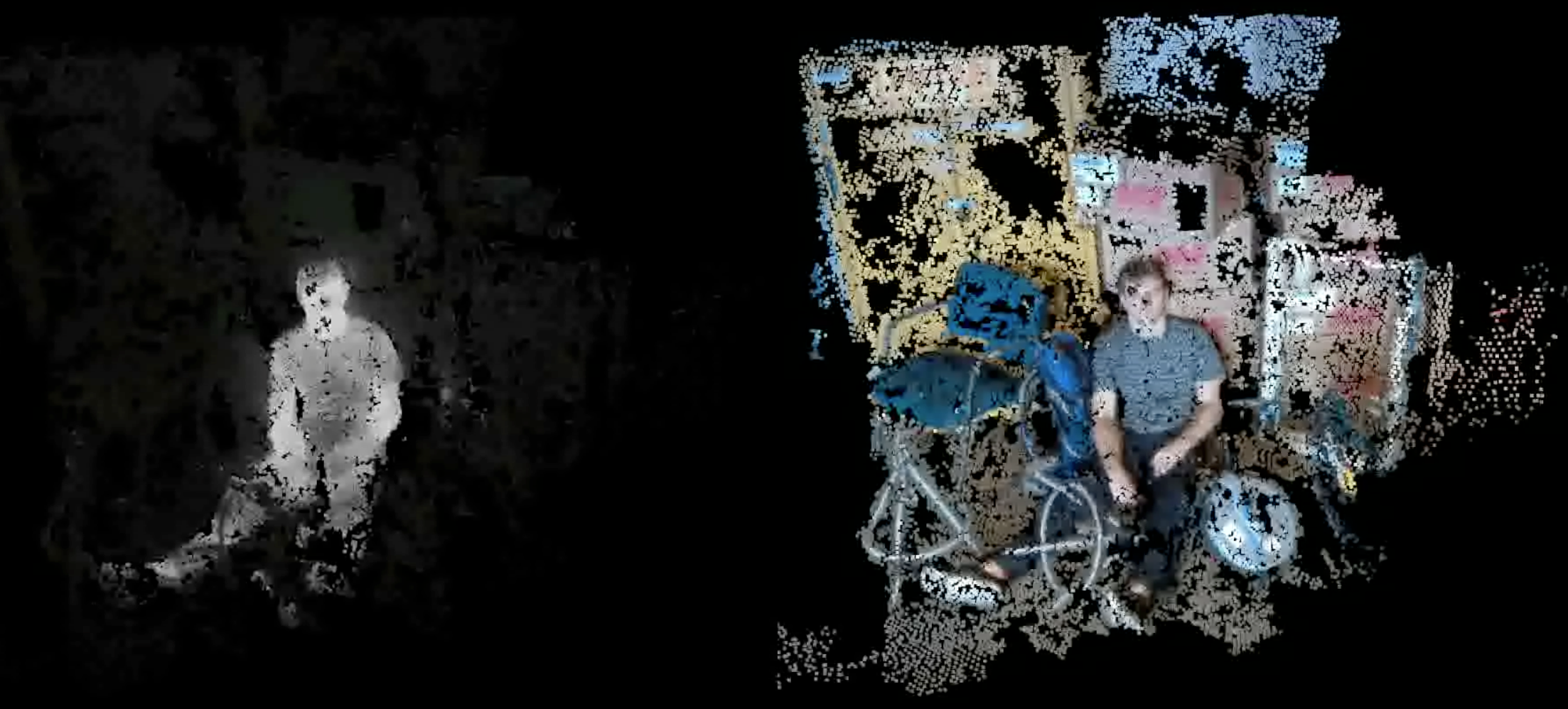}
    \caption{Victim sitting at a dark basement corner}
  \label{fig:garage}
\end{figure}

\begin{figure}
  \includegraphics[width=\linewidth]{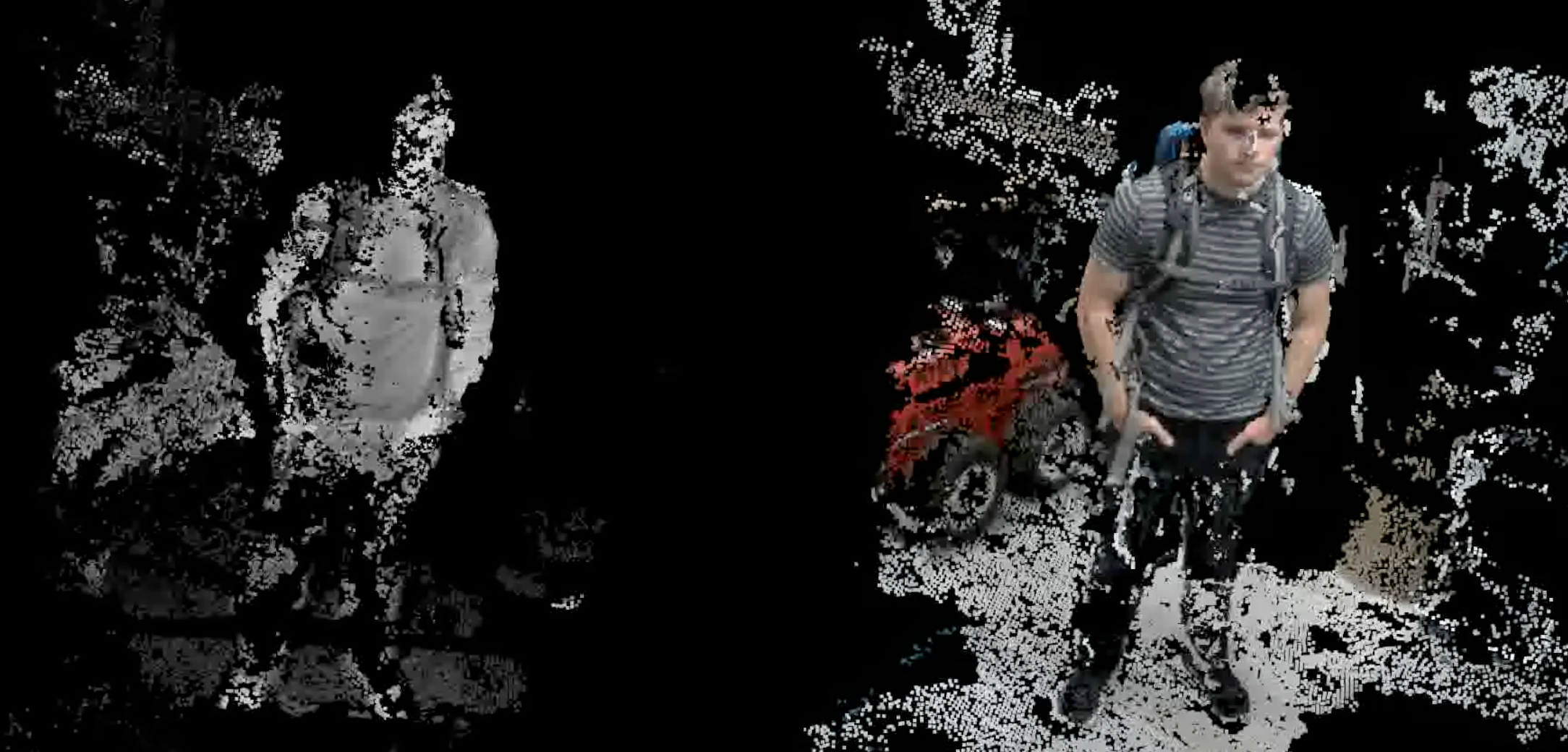}
   \caption{Person standing in front of a vehicle in lab}
  \label{fig:body}
\end{figure}

\begin{figure}
  \includegraphics[width=\linewidth]{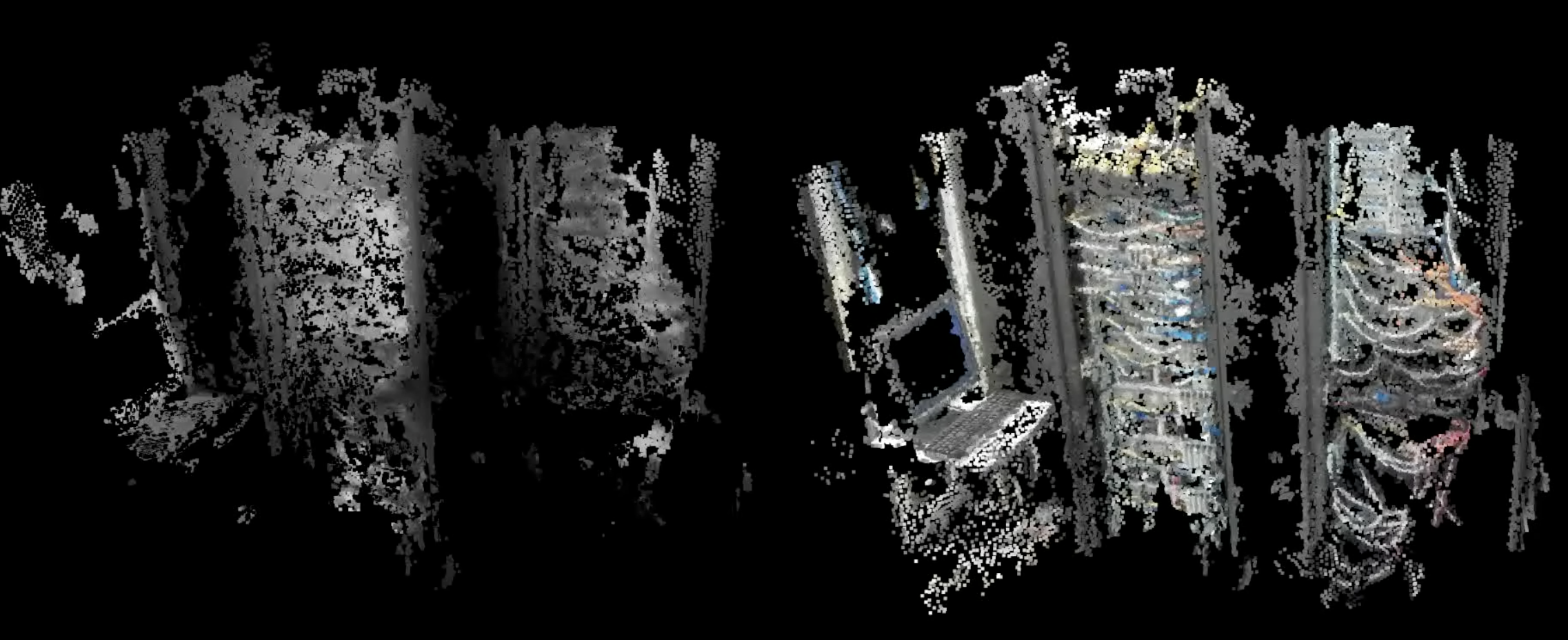}
   \caption{Hot server racks}
  \label{fig:server}
\end{figure}
\end{center}

\bibliographystyle{plainnat}
\bibliography{thermal}

\end{document}